%% file: acl_latex.tex
\pdfoutput=1
\documentclass[11pt]{article}

\usepackage[final]{acl}

\usepackage{times}
\usepackage{latexsym}

\usepackage[T1]{fontenc}

\usepackage{balance} % for balancing columns on the final page
\usepackage{graphicx} % for pdf, bitmapped graphics files
\usepackage{epsfig} % for postscript graphics files
\usepackage{amsmath} % assumes amsmath package installed
\usepackage{algorithm}
\usepackage{algpseudocode}
\usepackage{booktabs}
\usepackage{multirow}
\usepackage{amsfonts}
\usepackage[title]{appendix}
\usepackage[utf8]{inputenc}

\usepackage{microtype}

\usepackage{inconsolata}

\usepackage{tikz}
\def\checkmark{\tikz\fill[scale=0.4](0,.35) -- (.25,0) -- (1,.7) -- (.25,.15) -- cycle;}

\newcommand{\DONE}[1]{\noindent \textcolor{green}{\textbf{DONE}}\\ }

\title{Spatially-Aware Speaker for Vision-and-Language Navigation Instruction Generation}

\author{Muraleekrishna Gopinathan, Martin Masek, Jumana Abu-Khalaf, David Suter\\
{Centre for Artificial Intelligence and Machine Learning} \\
 {Edith Cowan University, 270 Joondalup Dr, Joondalup, WA 6027, Australia} \\
  {\small \texttt{\{k.gopinathan,m.masek,j.abukhalaf,d.suter\}@ecu.edu.au}}\\
  }

\begin{document}
\maketitle
\begin{abstract}
Embodied AI aims to develop robots that can \textit{understand} and execute human language instructions, as well as communicate in natural languages. On this front, we study the task of generating highly detailed navigational instructions for the embodied robots to follow. Although recent studies have demonstrated significant leaps in the generation of step-by-step instructions from sequences of images, the generated instructions lack variety in terms of their referral to objects and landmarks. Existing speaker models learn strategies to evade the evaluation metrics and obtain higher scores even for low-quality sentences. In this work, we propose SAS (Spatially-Aware Speaker), an instruction generator or \textit{Speaker} model that utilises both structural and semantic knowledge of the environment to produce richer instructions. For training, we employ a reward learning method in an adversarial setting to avoid systematic bias introduced by language evaluation metrics. Empirically, our method outperforms existing instruction generation models, evaluated using standard metrics. Our code is available at \url{https://github.com/gmuraleekrishna/SAS}.
\end{abstract}

\section{Introduction}

Incorporating language understanding in robots has been a long-standing goal of the NLP and robotic research community. Specifically, the Vision-Language Navigation (VLN) task requires robots to follow natural language instructions grounded on vision to navigate in human living spaces. Although humans generally follow navigational instructions well, training robots to follow natural language instructions remains a challenging problem. Detailed navigation instructions may include landmarks, actions, and destinations. Recent work has succeeded in improving instruction understanding of robots by augmenting instruction and trajectory training data \cite{Hong2020FGR2R,Wang2022CCC,Wang2023LANA}. They showed that using machine-generated instructions from a large number of navigational paths sampled from real houses helps robots navigate successfully even in previously unseen environments. However, there is still room for improvement, as the quality of machine-generated instructions is clearly lower compared to human annotations \cite{Zhao2021Eval}.  

\begin{figure}[t!]
    \centering
    \includegraphics[width=\columnwidth]{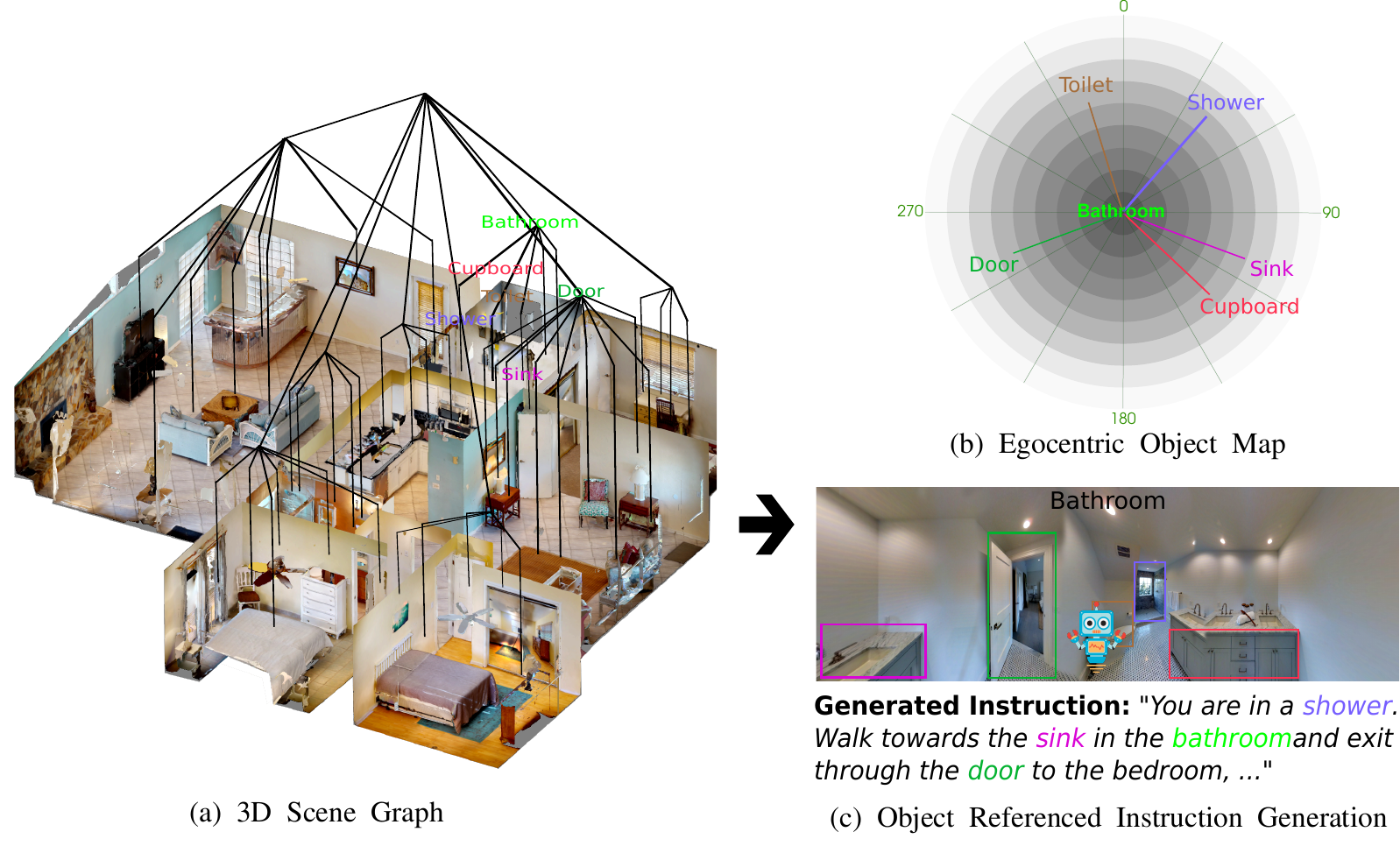}
    \caption{Extracting 3D scene relationships from house environments  (a,b) can improve instruction generation by including object references (c).}
    \label{fig:poster}
\end{figure}

In this work, we present a novel instruction generation model that can produce a variety of human-like instructions using semantic and structural cues from the environment. Our method uses rooms, interesting landmarks, objects, inter-object relations, object locations, and spatial features to produce richer instructions that can be used by robots and humans alike (Fig. \ref{fig:poster}). Our \underline{S}patially-\underline{A}ware \underline{S}peaker (SAS) model generates information-rich instructions by leveraging expert demonstrations that map trajectories to verbal directions. Incorporating spatial references within these instructions is critical, as they convey the environmental layout, highlight key landmarks relevant to the actions taken, and gauge the progression of navigation. At its core, SAS employs a sequence learning framework that is fine-tuned through a combination of adversarial learning rewards and multiple objectives aimed at enhancing its linguistic generation capabilities.

The architecture of SAS is based on an Encoder-Decoder model, which processes a sequence of viewpoints and corresponding actions that define a navigational path, subsequently generating a coherent set of instructions. During the encoding phase, the model extracts vital information from visual inputs, such as object categories (e.g., cupboard, bed), spatial relationships between objects (e.g., on top of, near, under), object placements, and significant landmarks within the viewpoint (e.g., bedroom, kitchen). These elements are combined with navigational actions to form a comprehensive vision-action representation that computes the temporal order.

The decoding phase acquires linguistic capabilities by linking this latent representation to the instructions encountered during training. An adversarial learning objective is introduced to encourage the generation of varied sentences, mitigating the potential biases that automatic evaluation metrics introduce. Through this novel approach, SAS outperforms existing instruction generation models on VLN datasets evaluated using standard language evaluation metrics.

Our contribution is as follows.

\begin{itemize}
    \item We introduce a novel speaker model (SAS) that can incorporate semantic and structural viewpoint features into the instruction.
    \item We develop an adversarial reward learning strategy, that rewards diverse instructions, to train our SAS model.
    \item We introduce a large scale silver dataset for automatic data augmentation. 

\end{itemize}

\input{relatedwork}

\input{method}

\input{experiments}

\bibliography{anthology,bibliography}

\input{appendix}
\end{document}

%% file: relatedwork.tex
\section{Related Work}
\subsection{Natural Language Navigational Guidance and Following}
Methods for modelling human and robot behaviours for the generation and execution of natural language instructions span several disciplines, including cognitive psychology \cite{Shawn1986}, sociology \cite{Andrew2000}, natural language processing (NLP) \cite{daniele2017navigational}, and robotics \cite{Wang2023LANA}. Studies show that adequate navigational instructions have \textit{directions}, \textit{landmarks}, \textit{region descriptions} and \textit{turn-by-turn actions} \cite{Look2005}. These instructions are also beneficial to human-machine interaction, particularly in embodied agents.

The embodied navigation problem has been receiving attention from multiple research domains such as robotics, NLP and scene understanding \cite{Anderson2018R2R,Dorbala2023}. Recent studies have shown that VLN agents learn better on machine-generated examples \cite{Fried2018SF,Tan2019EnvDrop,Wang2023LANA}. These methods, generally called \textit{Speaker models}, are still far from generating human-like instructions, as exhibited by their lower machine-generation evaluation scores. Our work aims to improve the quality of the generated instructions over baseline models by including landmarks, actions, and directions.
\subsection{Spatial and Semantic Scene Understanding for Embodied Navigation}
Neuroscience has shown that humans, like others in the animal kingdom, use spatial and temporal cues to build a cognitive map of their surroundings \cite{KUIPERS1978129}. These cognitive maps are crucial for manipulating or navigating the environment. Inspired by this, recent studies in robotics and embodied navigation have used vision models to infer the structure of the environment \cite{Kuo2023SEA,Gopinathan2021ISSU}, spatial relationships among objects in the scene \cite{Qi2020OAAM,Moudgil2021SOAT} and environment layout \cite{Gopinathan2023} to learn about the environment \cite{Song2022MTrack}. While these methods utilise structure, spatial and semantic knowledge in various combinations to learn vision-language association, they are not applied to instruction generation task. In this work, we use all four aspects of a satisfactory instruction - \textit{landmark} through visual encoding, \textit{directions} through directional encoding, \textit{turn-by-turn actions} using action encoding and \textit{region descriptions} as semantic encoding - to generate richer instructions.
\subsection{Reinforcement learning for Instruction Generation}
\label{sec:rl}
Reinforcement learning (RL) has been successful in machine generation tasks such as translation, instruction generation, captioning, and storytelling. In this paper, the primary objective is to maximise the expected return of a word-generating policy.  RL instruction generation methods are found to learn the target distribution better than traditional maximum likelihood estimation (MLE) algorithms due to the inherent \textit{exposure bias} \cite{Arora2022EB}. Applying RL to learn the target distribution requires extensive feature and reward engineering. Instead, inverse reinforcement learning (IRL) is proposed to infer the expert's reward function. Using an adversarial setting, IRL has been shown to improve visual story telling \cite{Wang2018NoMetrics}. 

Existing work in VLN have studied natural language instruction generation \cite{Fried2018SF, Wang2022CCC, Wang2019RCM} as a sequence generation problem, however, they focus on navigational success of the overall agent over instruction quality. Duo et al. \citet{Dou2022FOAM} optimise their \textit{Speaker} by using the similarity between itself and the gradient of the navigation model as a reward for RL. The authors evaluated their method on BLEU, a metric which does not guarantee high-quality instructions. \citet{Zhao2021Eval} discovered that in the context of dialogue generation and navigational tasks, the majority of n-gram-based automatic language evaluation metrics show a weak correlation with human-annotated instructions.

Inspired by these studies, we adopt an IRL-based reward learning strategy to produce high-quality navigation instructions by indirectly learning the reward from language metrics. This mitigates exposure bias and avoids the model from gaming the metrics to achieve high evaluation scores - even with low-quality instructions.

%% file: method.tex
\section{Problem Definition}
Here we present the task aimed at generating linguistic instructions from navigational demonstrations. The generation of instructions is posited as the converse operation to the standard VLN task, where an agent executes a navigation instruction in a house. For this, we develop a \textit{Speaker} agent that synthesises a coherent set of natural language instructions from a sequence of navigational actions along a trajectory.

At each discrete time step $t$, the \textit{Speaker} agent is presented with a panoramic visual observation $O_t$ and a directional action $a_t$ that signifies the transition to the next viewpoint within the trajectory. Upon completion of a navigational episode, the agent outputs the complete instruction of the traversed path, that is, $X = \{w_0,\ldots,w_l\}$. Formally, the objective of the \textit{Speaker} is to minimise the negative log-likelihood for ground-truth instruction \textbf{X} conditioned on trajectory $T = \{O_1,\ldots,O_N\}$ with parameter $\theta$:
\vspace{-0.5pt}
\begin{equation}
L_{\theta} = -\sum_\theta \log(p(X|T; \theta)))    
\end{equation}

\section{Preliminaries}
\subsection{Path Mixing}
\label{sec:path_mix}
\begin{figure}[htp]
    \centering
    \includegraphics[width=0.9\columnwidth]{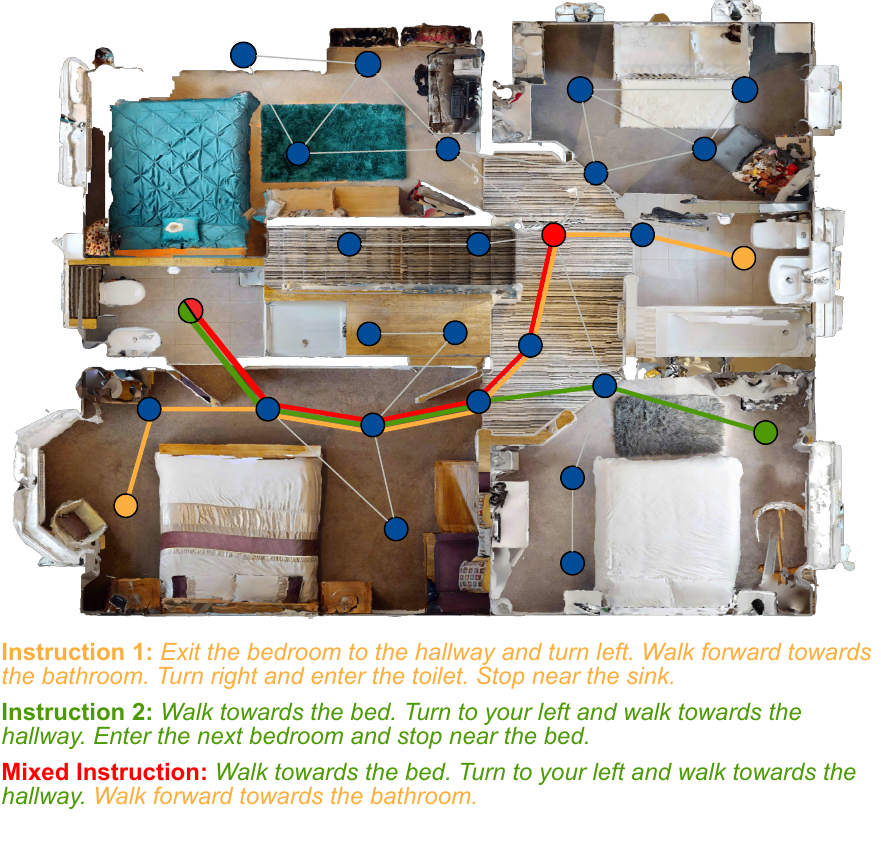}
    \caption{Path Mixing (PM) using fine-grained paths from R2R dataset. Original paths~\includegraphics[width=0.6em]{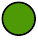}$\rightarrow$~\includegraphics[width=0.6em]{green.pdf} and ~\includegraphics[width=0.6em]{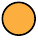}$\rightarrow$~\includegraphics[width=0.6em]{yellow.pdf} are mixed to generate ~\includegraphics[width=0.6em]{green.pdf}$\rightarrow$~\includegraphics[width=0.6em]{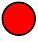}.}
    \label{fig:pathmix}
\end{figure}
Prior work in VLN have shown that more instruction examples can improve an agent's performance in previously unseen environments \cite{Moudgil2021SOAT,Wang2022CCC}. Hence, to supplement training data, we mix parts of trajectories from the FGR2R dataset \cite{Hong2020FGR2R} (which is derived from the R2R \cite{Anderson2018R2R} dataset) to obtain additional instruction-trajectory pairs. 

R2R dataset contains paths and human-annotated instructions for navigating inside 3D scanned house environments from the Matterport3D (MP3D) dataset \cite{Chang2017MP3D}. Further fine-grained (turn-by-turn) or \textit{micro-instructions} are available in the FGR2R dataset. We adopt FGR2R to enhance our training data by algorithmically combining parts of its trajectories, creating new instruction-trajectory pairs. Unlike REM \cite{Moudgil2021SOAT,Liu2021REM}, which randomly mixes data from different houses, we only mix trajectories from the same house to ensure visual and object referral consistency.

First, we identify key edges in the graph to mix trajectories, focusing on start edges $\varepsilon_{start}$ and end edges $\varepsilon_{end}$ of navigation paths. These edges are crucial as they relate to start and end of instructions (e.g. "Walk away from the desk, Turn right"). Edges with micro-instructions\footnote{Part of the instruction pertinent to one edge of the path} that do not contain a \verb|NOUN|, \verb|VERB| are ignored to avoid partial or inconsequential (\textit{wait there}) actions. Then, we mix the remaining transition edges $\varepsilon_{trans}$ to form a trajectory from $\varepsilon_{start}$ to $\varepsilon_{end}$. Nodes that are too close to each other, leading to a lack of visual diversity or repetitive micro-instructions are also avoided. We connect edges based on the following criteria: (1) the distance between any two nodes (except the start and end nodes) should not exceed 3m, (2) the angles between edges should prevent looping; (3) the start and end nodes should not share an edge; (4) micro-instructions from shared edges of different trajectories are selected randomly.

The final instruction combines these micro-instructions, and the trajectory is a sequence of edges (as shown in Fig. \ref{fig:pathmix}). Using this method, we generated 162k instruction-trajectory pairs with path lengths ranging from 5m to 20m. The dataset\footnote{Available at \url{https://zenodo.org/records/10396782}} averages 7.27 views per trajectory, a mean path length of 14.4m, and about 82 words per instruction. More dataset statistics are added to the supplementary material.

\subsection{Action Parsing}
\label{sec:action_parsing}
To allow the speaker to learn action phrases by associating them to the navigational actions, we automatically extract the action phrases from the instruction for training. For this, we use spaCy's \cite{spaCy2020} dependency parsing and part-of-speech tagging to identify verb forms that are transitive, indirect transitive, direct transitive and also the part-of-speech forms such as nouns, adverbs, adpositions, interjections and determiners (refer Appendix \ref{alg:action_parsing} for more details). We use this algorithm to identify action phrases from each step of the instruction and classify parts of sentences to either actions or other phrases. The identified micro-instructions are used to train the decoder part of SAS ($\S\ref{sec:atgl}$).

\section{Spatially-Aware Speaker (SAS) Model}
\label{sec:sas-method}
In this section, we present our Spatially-Aware Speaker model. SAS is an encoder-decoder model which generates an instruction for a trajectory when the sequence of viewpoints and actions are provided. The \textit{trajectory encoder} produces visual-action context from the trajectory viewpoints and the corresponding actions. This context is used by \textit{instruction decoder} to generate instruction.

To incorporate spatial and semantic awareness, we provide three crucial pieces of information to the model, namely: Action Encoding, Structural Encoding and Semantic Encoding. These are explained in the following sections.  
\paragraph{Action Encoding}
\label{sec:action_encoding}
The action taken between viewpoints in a trajectory is represented by Action Encoding. The visual action encoding is the current heading and elevation of the agent with respect to the next view direction. The relative elevation ($\theta$) and heading ($\phi$) angles are encoded as $E_a = [\cos\theta, \sin\theta, \cos\phi, \sin\phi]$. 
% Additionally, to reinforce vision-language-action association, we first identify the action phrases from the ground-truth instructions. The unique action phrases or \textit{action classes} are aligned to the sub-instructions using soft attention between the visual-language context and the embedded action phrases in the decoder. This forces the decoder to attend to the action phrases from the instruction and the visual cues from the scene.
\paragraph{Structural Encoding}
Structural encoding provides knowledge of the egocentric locations of objects with respect to the \textit{Speaker}. In the panoramic view of each viewpoint, we extract an object's location in the image frame, as well as its size and distance to the agent in order to represent a complete pose of the object relative to the agent. The image frame location is obtained from the location of the object's bounding box detected by a Faster R-CNN \cite{Ren2015FasterRCNN} detector trained on the Visual Genome \cite{krishna2017visual} dataset. The size and distance of the object are obtained by projecting (inverse pinhole camera projection) the bounding box to the point cloud and measuring the centroid and volume of the contained point cloud. This provides an estimate of the size and distance to form the object descriptions. Effectively, the structural encoding is a combination of object features $f_o$, object location $(c_x, cy)$, size $s_o$ and distance $d_o$ , respectively, i.e. $E_{so} = [f_o, (c_x, cy), s_o, d_o]$.  
\paragraph{Semantic Encoding}
To provide inter-object relationships we use 1600 object classes FasterRCNN and relationships extracted from ConceptNet \cite{speer2017conceptnet}. The object-to-object-room semantic features are a combination of GloVe embedding $G$ of the respective token,  $e^{obj}_{i,j,k}$=
\texttt{\{G(<$obj_i$>);G(<$rel_j$>);G(<$obj_k$>)\}}. 

Furthermore, we encode the relationship from room to object using the \textit{in the} relation as $e^{room}_{l,m}$=\texttt{\{G(<$obj_l$>);G(<$in$>);G(<$room_m$>)\}}.
For each viewpoint, we encode one $e_{room}$  and one $e^{obj}$ per view direction with the highest detection confidence. In effect, we obtain the semantic encoding per viewpoint, $E_{sm} = \{(e^{obj}_1,e^{room}),\ldots, (e^{obj}_{36},e^{room})\}$.  
\paragraph{Panoramic Room-Object Attention}
\label{sec:psa_attention}
Finally, we combine structural and semantic knowledge with panoramic view to obtain a panoramic knowledge feature. For this we first concatenate the candidate feature $f_c$ with the Structural Encoding $E_{so}$ and Semantic Encoding $E_{sm}$:

\begin{equation}
    f_{cs}  =  \textbf{W}[f_c;E_{sm};E_{so}]
\end{equation}
where $\textbf{W}$ is the trainable projection. We apply an attention module to make the model attend to information from different sub-spaces. The two projections $Q$(query) and $K$(key), which are from the action embedding and the candidate-semantic embedding,  respectively, are applied to the attention as:
\begin{eqnarray}
    \alpha_k & = & \mathrm{softmax}\Bigg(\frac{Q(h^a_t) K^T(f_{cs})}{\sqrt{D_k}}\Bigg)\\
    c_k & = & \sum_{i,j} \alpha_{i,j}  f_{p_{i,j}}\\
    g_t & = & \mathrm{tanh}(\textbf{W}[c_k; h^a_t])
\end{eqnarray}

where $c_\alpha$ is the context vector, $D_k$ is the hidden size of the attention layer, $h^a_t$ is the hidden action state of the \textit{ trajectory encoder} and $g_t$ is the gated output. The affinity matrix $f_p$ governs the information flow between neighbouring view patches of the panorama. Finally, we capture the panoramic room-object feature using an LSTM:
\vspace{-0.5pt}
\begin{equation}
    h_t^v  =  \mathrm{LSTM}_v(g_t, h^v_{t-1}), \forall{t}={1,\ldots,N}
\end{equation}

\subsection{Trajectory Encoder}
\label{sec:trajectory_encoder}
The trajectory encoder consists of a multilayer bidirectional LSTM to summarise the input sequence at each time step conditioned on the navigational trajectory. This bidirectional approach ensures action context at each step is influenced by both the historic and the future actions in the sequence. 

The first $\mathrm{BiLSTM}_A$ encodes navigation actions from the ground truth action $a$. Scaled-dot attention is applied to the action hidden state $h^a$ and the panoramic visual features $f_v$ giving a context vector $c_w$. A second LSTM encodes the change of the context vector as $h_t^{v,a}$. This hidden state is used by the decoder to learn visual and language alignment. Formally:
% \begin{equation}
% \end{equation}
\begin{eqnarray}
        h^a_t & = & \mathrm{BiLSTM}_A(a_t, h^a_{t-1})\\
    \alpha_w & = & \mathrm{softmax}\Bigg( \frac{Q_a(h^a_t)K_v^T(h^v_t)}{\sqrt{D_k}}\Bigg)\\
    c_w & = & \sum_{i,j} \alpha_w h^a_t\\
    \Tilde{h} & = & \mathrm{tanh}(\textbf{W}[c_w;h^v_t])\\
    \hat{h}_t^{v,a} & = & \mathrm{BiLSTM}_{VA}(c_w, \Tilde{h})
\end{eqnarray}
where $a$ is the action embedding, $h$ is the hidden state. Also, $Q_a = F^a_q (h_t^a)$, $K_v = F^v_k(f_v)$, and $D_k$ are query and key vectors, hidden dimension size of the soft attention, respectively.

\subsection{Instruction Decoder}
\label{sec:instruction_decoder}
Our instruction decoder is guided by semantic and structural knowledge from the environment. The basic structure of the decoder is as follows. When the decoder is provided with the previous instruction token and the visual-action context $h_t^{v,a}$, it applies an LSTM to encode the instruction token embedding from the previous time step:
\vspace{-0.5pt}
\begin{eqnarray}
    w_{t-1}^{emb} = \mathrm{embedding}(w_{t-1})\\
    h^X_t = \mathrm{LSTM}_X(w_{t-1}^{emb}, h_{t-1}^{v,a})
\end{eqnarray}
we apply a scaled-dot attention on the projected instruction context $Q_X = F_q(h_t^X)$ and the vision-action context $K_{va} = h_t^{v,a}$ and $V = F_v(h^X_t)$:
\begin{equation}
    \hat{h}_{t}^{vaX}  =  \mathrm{Attention}(Q_X,K_{va},V_X)
\end{equation}
Finally, the next predicted word is the token of maximum probability:
\begin{equation}
    \label{eq:word_pred}
    w_{t} =  \mathrm{arg} \max (\textbf{W}\hat{h}_{t}^{vaX})
\end{equation}

% The probability of generating the k-th word token at time step t is softmax over a linear transformation of the attentive hidden ˆht. The loss Lt is the negative log likelihood of the ground truth word token. 
\section{Training}
\label{sec:training}
The Encoder (\S\ref{sec:trajectory_encoder}) and Decoder (\S\ref{sec:instruction_decoder}) modules of SAS are trained end-to-end. SAS model predicts the next token $w_t$ based on the complete trajectory $T = \{O_1,\ldots,O_N\}$ and all previous tokens $w_{<t}$. The trajectory embedding from the encoder and $w_{<t}$ is fed to the decoder to produce a probability distribution $p_L$ over the next word token. This distribution is sampled as in \eqref{eq:word_pred} to predict the next token $w_t$.

SAS model is trained using a mixture of a Teacher-Forcing (TF) \citep{Williams1989ALA} and the ARL method (\S\ref{sec:trainingobjectives}). In TF, the decoder generates the next token based on a ground truth token instead of using its predicted token as in the Student Forcing (SF) strategy. This method has been shown to improve the baseline methods \cite{Anderson2018R2R}. Next, we describe reward learning in detail.

\subsection{Reward Learning}
\label{sec:trainingobjectives}
\begin{figure}[]
    \centering
    \includegraphics[width=\columnwidth,trim={1pt 4pt 1pt 5pt},clip]{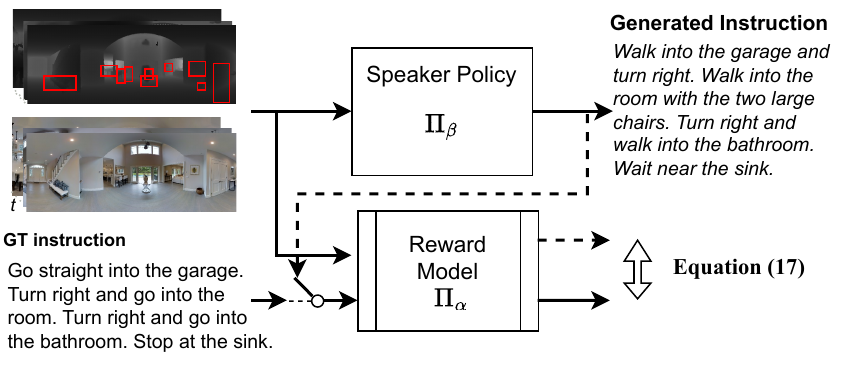}
    \caption{Adversarial training of SAS model. SAS learns to generate instruction, while reward model learns the reward function from ground truth data. The learned reward function is employed to optimise the policy}
    % \Description{Generator Adversarial Network contains our Speaker model as policy and a Reward model to predict the reward for generating good instructions}
    \label{fig:gen-disc}
\end{figure}
Applying Reinforcement Learning (RL) for an instruction generation task using automatic evaluation metrics as reward functions, causes the model to \textit{game} the metrics. Instead, in reward learning, a reward model learns the best reward for human-annotated and speaker-generated instructions.
In this strategy, we use a Generative Adversarial Network (GAN) \citep{Goodfellow2014GAN} architecture - with a Policy and Reward model - to learn an association between the instruction and reward distributions (Fig. \ref{fig:gen-disc}). In an RL sense, this is a Markov decision process (MDP) with SAS as a policy $G_\beta(\cdot)$, generating words $X^G$ and receiving a \textit{reward} $r(X^G)$ score. The objective is to maximise the expected reward of $G_\beta$, $\mathbb{E}_{X^G \sim G_\beta}[r(X^G)]$. In reward learning, the reward distribution is learned from the demonstrations, rather than adopting a function to provide a reward.

We adapt the idea of Reward-Boltzmann distribution from \citet{Wang2018NoMetrics} to approximate a reward obtained for identifying \textit{fake} (speaker-generated) or \textit{real} (human-annotated) instructions. The approximate reward distribution $\Pi_\alpha$ for an instruction $X$ is defined as: 
\begin{eqnarray}
\label{eq:rewardboltsmann}
\Pi_\alpha(X) & = & \frac{e^{R_\alpha(X)}}{\underset{{X}}{\sum} e^{R_\alpha(X)}}    
\end{eqnarray}
where $R_\alpha$ is the reward function. The optimal reward function is achieved when \eqref{eq:rewardboltsmann} is equal to the distribution of human-annotated instructions. We optimise this using an adversarial two player min-max game between (1) $\Pi_\alpha$ maximising its similarity (measured by KL-divergence) with the empirical distribution $\Pi_\epsilon(X)$ of the training dataset and minimising its similarity with the distribution of \textit{fake} instructions from the policy $\Pi_\beta$ and (2)  maximising the similarity of the policy distribution $\Pi_\beta$ with that of $\Pi_\alpha$. Formally, the objective is:
\vspace{-0.5pt}
\begin{equation}
    \max_{\beta} \min_\alpha KL(\Pi_\epsilon||\Pi_\alpha) - KL(\Pi_\beta||\Pi_\alpha)
\end{equation}
This is optimised through a policy-gradient-based reinforcement learning method.
\paragraph{Reward Model}
We investigate two reward models $R_\omega$ based on the CNN and RNN models. The CNN-based discriminator uses the GloVe embedded instruction $\textbf{X}_{emb}$ and visual feature $Oi$ to produce a reward score $R_\omega$ after the activation function.  Formally, the CNN and RNN rewards are, respectively:
\vspace{-0.5pt}
\begin{eqnarray}
       R^{CNN}_\omega = \textbf{W}_R(\mathrm{Conv}(X_{emb}); \textbf{W}_O O_i) \\
        R^{RNN}_\omega   = \textbf{W}_R(X_{emb}; \textbf{W}_O \mathrm{AvgPool}(O_i))
\end{eqnarray}

where $W_R$ and $W_O$ are linear learnable weights, $Conv$ represents the convolution layers followed by the mean pooling operation, and $[;]$ is the feature concatenation. The final sigmoid activation is not shown for brevity.

Both the Speaker policy and the reward models are trained alternately using the Adversarial Reward Learning (ARL) algorithm (Algorithm \ref{alg:arl}). Reward models are evaluated in our ablation studies (\S\ref{sec:ablation_study}).

\subsection{Supervised Learning}

\begin{table*}[htp!]
    \centering
    \caption{Benchmarking results of Speaker-based models ($\S\ref{sec:experiments}$) on R2R dataset}
    \resizebox{\textwidth}{!}{%
    \begin{tabular}{lcccccccccc}
       \toprule
    \multicolumn{1}{c}{\multirow{2}{*}{Methods}}  & \multicolumn{5}{c}{R2R ValSeen} & \multicolumn{5}{c}{R2R ValUnseen} \\
 \cmidrule(lr){2-6}  \cmidrule(lr){7-11}
 & \multicolumn{1}{c}{SPICE} & \multicolumn{1}{c}{CIDEr} &  \multicolumn{1}{c}{METEOR}  & \multicolumn{1}{c}{ROUGE} &
 \multicolumn{1}{c}{BLEU-4}
 & \multicolumn{1}{c}{SPICE} & \multicolumn{1}{c}{CIDEr} &  \multicolumn{1}{c}{METEOR}  & \multicolumn{1}{c}{ROUGE} &
 \multicolumn{1}{c}{BLEU-4}\\ 
 \midrule
    Speaker-Follower \cite{Fried2018SF} & 22.1 & 43.7 & 23.0 & 49.5 & 28.3 & 18.9 & 37.9 & 21.7 & 48.0 & 26.3\\
    % LandmarkSelect [2] &19.7 & 13.2 &23.1 & 35.7
    EnvDrop \cite{Tan2019EnvDrop} &24.3 & 47.8&24.5 & 49.6 & 27.7& 21.8 & 41.7 & 23.6 & 49.0 & 27.1 \\
    CCC \cite{Wang2022CCC} & 23.1&\textbf{54.3}&23.6&49.3& 28.7&21.4 & \textbf{46.1} & 23.1 & 47.7 & 27.2\\
    % KEFA\cite{Zeng2023KEFA} & 27.2&52.3&25.4&51.8& 32.6&23.4 &  42.7 & 24.4 & 50.3&28.3 \\
    LANA \cite{Wang2023LANA} & 25.6 & 53.3 & 24.5 & 50.3 & 31.4& 22.6 & 45.7 & 23.8 & 49.8&29.8 \\ \hline 
    $\textsc{SAS}_{TF}$ (Ours) &27.9&53.1 &28.3 &54.9 & 30.2& 22.2 & 44.9 & \textbf{26.3} & 55.4 & 30.2\\
    $\textsc{SAS}_{ARL+TF}$ (Ours) &\textbf{28.1}&51.6 &\textbf{29.7} &\textbf{56.8} & \textbf{31.4}& \textbf{24.8} & 43.5 & 25.7 & \textbf{56.5} & \textbf{33.8}\\
 \bottomrule
    \end{tabular}%
    }
        \label{tab:r2rinstruction}

\end{table*}

\begin{table*}[htp!]
    \centering
    \caption{Benchmarking results of Speaker-based models ($\S\ref{sec:experiments}$) on R4R dataset}
    \resizebox{\textwidth}{!}{%
    \begin{tabular}{lcccccccc}
    \toprule
    \multicolumn{1}{c}{\multirow{2}{*}{Methods}}  & \multicolumn{4}{c}{R4R ValSeen} & \multicolumn{4}{c}{R4R ValUnseen} \\
    \cmidrule(lr){2-5}\cmidrule(lr){6-9}
 & \multicolumn{1}{c}{SPICE} & \multicolumn{1}{c}{CIDEr} &  \multicolumn{1}{c}{METEOR}  & \multicolumn{1}{c}{ROUGE} & \multicolumn{1}{c}{SPICE} & \multicolumn{1}{c}{CIDEr} &  \multicolumn{1}{c}{METEOR}  & \multicolumn{1}{c}{ROUGE} \\ 
 \midrule
    Speaker-Follower \cite{Fried2018SF} & 16.4 & 9.9 & 21.3 & 45.3 & 20.7 & 13.9 & 17.2 & 35.9 \\
    % LandmarkSelect [2] &19.7 & 13.2 &23.1 & 35.7
    EnvDrop \cite{Tan2019EnvDrop} & 20.9 & 21.6 & 24.5 & 47.3 & 21.8 & 20.0 & 18.7 & 36.3\\
    CCC \cite{Wang2022CCC} & 21.9 & 24.5 & 25.2 & 48.0 & 23.3 & 20.6 & 19.3 & 36.5 \\

    LANA \cite{Wang2023LANA} & 24.5 & \textbf{28.7} &  26.1 & 48.4 & 
    26.2 & \textbf{23.1} & 20.0 & 37.6
    \\ \hline 
    $\textsc{SAS}_{TF}$ (ours) &25.8 &26.5 &27.6 &50.1 & 27.4 & 21.7&21.1 & 38.4 \\
    $\textsc{SAS}_{ARL+TF}$ (ours) &\textbf{26.2} &22.3 &\textbf{28.9} & \textbf{50.3} & \textbf{28.1} & 22.5 & \textbf{22.8} & \textbf{39.2} \\
 \bottomrule
    \end{tabular}%
    }
    \label{tab:r4rinstruction}
\end{table*}

The supervised learning objectives used for both the teacher-forcing (used in the final model) and the student-forcing (for the ablation study) strategies are as follows.
\paragraph{Language modelling}
Conditioned on path $T$ and linguistic embedding $w_{<t}$, the probability of decoded words is optimised as a maximum likelihood estimation (MLE) problem:
\vspace{-0.5pt}
\begin{equation}
\label{eq:langmodelloss}
    \mathcal{L}_{LM} =  -\sum_{t \in \{1:N\}} \log(p_L(w_t|w_{<t}, T))
\end{equation}
where $p_L$ is the likelihood of a token given trajectory $T$ and ground truth instruction tokens $w^{GT}_{<t}$.
\paragraph{Unlikelihood training}
Even low-perplexity machine generation models are prone to repeating tokens when presented with small examples \cite{Holtzman2020}. To mitigate this, we apply Sequence-Level unlikelihood loss \cite{Welleck2020UL} on the decoded instruction that penalises repetition of word tokens $w$. The objective is to minimise the logarithmic likelihood of negative candidates (repeated tokens) $\mathcal{C}^t$ conditioned on previous tokens $w_{<t}$:
\begin{equation}
\label{eq:ulsloss}
    \mathcal{L}_{ULS} =  -\sum_{c\in\mathcal{C}^t} \log(1 - p_\mu(c| w_{<t}))
\end{equation}

% \subsection{Action-Phrase Order Loss}
% To ensure the action phrases in the generated instruction correspond to the ground truth instruction, we compute a cross-entropy loss between the locations of action phrases from ground truth instruction and locations of action phrases from generated instruction. 
\paragraph{Temporal Alignment Loss}
\label{sec:atgl}
We introduce a temporal alignment loss (TAL) to train the decoder to attend between action phrases and visual-action context. The decoder's attention matrix $A_{D}$, which represents attention between word tokens and the panoramic action context, is compared to the ground truth vision language alignment scores (\S\ref{sec:action_parsing}). Formally,
\vspace{-0.5pt}
\begin{equation}
    A_{GT} = 
    \begin{cases}
    1,& w \leftrightarrow o_i ;  w \in X, o_i \in O \\
    0,              & \text{otherwise}
    \end{cases}
\end{equation}

where $\leftrightarrow$ denotes action phrase-viewpoint alignment. $\mathcal{L}_{TAL}$ is the binary cross-entropy loss between $A_{GT}$ and $A_{D}$.

\paragraph{Total Objective}
The total training objective is the weighted sum of all losses, that is, $\mathcal{L} = \lambda_{LM}\mathcal{L}_{LM} + \lambda_{ULS}\mathcal{L}_{ULS} + \lambda_{TAL}\mathcal{L}_{TAL}$.

% Here we introduce a contrastive loss that optimizes attention applied during sentence decoding for same trajectory and different teacher instructions are similar while the attention applied for trajectories with some common sub-paths are different. Hence 3 pairs of instructions and trajectories are used; anchor instruction and trajectory pair ($X_{ref}$, $O_{ref}$), positive pair ($X_{pos}$, $O_{pos}$) where the paths share same sub-paths between them, and negative pair ($X_{neg}$, $O_{neg}$) where the paths do not share any common edges. We compute alignment using Dynamic Time Warping (DTW) algorithm with action phrases obtained from \ref{alg:action_parsing}. This ensures each decoded action phrases are aligned with the actions performed on the environment. In contrast to \cite{Wang2022CCC} we do not apply contrastive loss to the generated instruction. Instead we contrast the attention applied by the decoder for positive and negative instruction pairs. Let alignment $A_{pos}$ = $a_{pos}^Ta_{ref}$ for attention scores of positive and reference pairs respectively. Similarly, $A_{neg}$ = $a_{neg}^Ta_{ref}$ is the alignment for attention scores from negative and reference examples. The contrastive loss is defined as,

% \begin{equation}
%     \mathcal{L}_{CE} = -\log\Bigg[\frac{exp(A_{pos}/\tau)}{\sum exp(A_{neg}/\tau) + exp(A_{pose}/\tau)}\Bigg]  
% \end{equation}

%% file: experiments.tex
\section{Experiments}

\label{sec:experiments}
\subsection{Datasets}
Our method is assessed using two datasets from the Vision-and-Language Navigation (VLN) field: R2R, which features brief trajectory paths and instructions for locating rooms, and R4R, an extension of R2R that links two adjacent tail-to-head trajectories along with their associated instructions to produce longer instructions. For training, we augment R2R dataset with the Path Mixing (PM) dataset $\S\ref{sec:path_mix}$. 

%The PM dataset has an average of 7.27 views per path, 14.4m average trajectory length and 82 words per instruction.

\subsection{Evaluation Metrics}
We evaluate the performance of SAS instruction generation using standard language metrics such as SPICE \cite{Anderson2016Spice}, CIDEr \cite{Vedantam2015Cider}, ROUGE \cite{lin-2004-rouge}, METEOR \cite{denkowski2014meteor} and BLEU-4 \cite{Papineni2001BLEU}. SPICE is considered the main metric in navigational instruction generation tasks \cite{Zhao2021Eval, Wang2022CCC}.  A high SPICE score indicates high lexical and semantic similarities of sentences and higher success of an embodied agent, which is important for navigational instructions \cite{Zhao2021Eval}.
\subsection{Implementation Details}
We use Speaker-Follower \cite{Fried2018SF}, a popular baseline used in previous work, for our experiments. The model is trained for 100k epochs ($\approx$14h) using NVIDIA RTX A6000 with batch size 8 and AdamW \cite{LoshchilovH19} optimiser with learning rate 5e-4. The visual feature and GloVe embedding sizes are 2048 and 300, respectively. The hidden size for the attention layers $D_k$ is 512. The training objective weights are set in the ratio $\lambda_{LM}:\lambda_{ULS}:\lambda_{TAL} = 2:1:1$ to prioritize language learning over repetition and alignment.. The hidden dimensions for the two-layer $\mathrm{BiLSTM}$ and the QKV sizes in $\mathrm{Attention}$ are set to 768d. We report the evaluation values of a single run.  

\section{Results}
 We evaluate two variations of SAS to measure the effectiveness of the proposed method (1) SAS with teacher-forced $SAS_{TF}$ using PM augmentation and temporal alignment (TAL) and (2) $SAS_{ARL+TF}$ a mixture of TF and ARL using all augmentation and supervised learning objectives. From the results (Table \ref{tab:r2rinstruction}), we see that our SAS method improves on the baseline by a large margin. Reward learning has absolute improvements of +5.9 (SPICE), +5.6 (CIDEr), +4 (METEOR), +8.5 (ROUGE), and +7.5 (BLEU-4) in the R2R ValUnseen split compared to the Speaker-Follower baseline. In the long instruction dataset R4R (Table \ref{tab:r4rinstruction}), both variations show better scores on all metrics (SPICE: +7.4, METEOR: +8.6, ROGUE: +3.3 and CIDEr +8.6). In both \texttt{ValUnseen} splits, the CIDEr score is markedly impacted (R2R: -2.6 and R4R: -0.6)  when comparing the overall best model to the baseline models.
\subsection{Discussion}
SAS shows a better instruction generation capability with TAL, which goes to show that supervising the speaker with action-only sentences is useful. In addition, our method outperforms the baseline in most of the major metrics, specifically in SPICE. This shows that spatial awareness is beneficial for Speaker models.  In the R2R and R4R datasets, both $SAS_{TF}$ and $SAS_{ARL+TF}$ have the best scores compared to the previous models, except for CIDEr. CIDEr rewards lexical similarity over semantic similarity. As temporal alignment of actions is pertinent in instruction generation and not in the lexical order, lower scores in these metrics definitely do not reflect wrong actions. Among these metrics, a higher SPICE score shows that our model generates temporally consistent instructions. \citet{Zhao2021Eval} observe that SPICE correlates with human way-finding performance, VLN agent navigation performance, and subjective human judgements of instruction quality, when averaged over many instructions. This correlation is not observed at the instruction level due to the high variance between the words used in the instructions. Human evaluation of generated instructions should be performed to ensure the actual quality of the instruction. In our study, we consider the high SPICE score as an early indicator of robust pathfinding performance for both agents and humans. It also reflects that human judgement of the quality of SAS-generated instructions is also high. 

\subsection{Qualitative Results}
\begin{figure}[]
    \centering
    %  [trim={left bottom right top},clip]
    \includegraphics[width=0.9\columnwidth,trim={1pt 8pt, 1pt 12pt},clip]{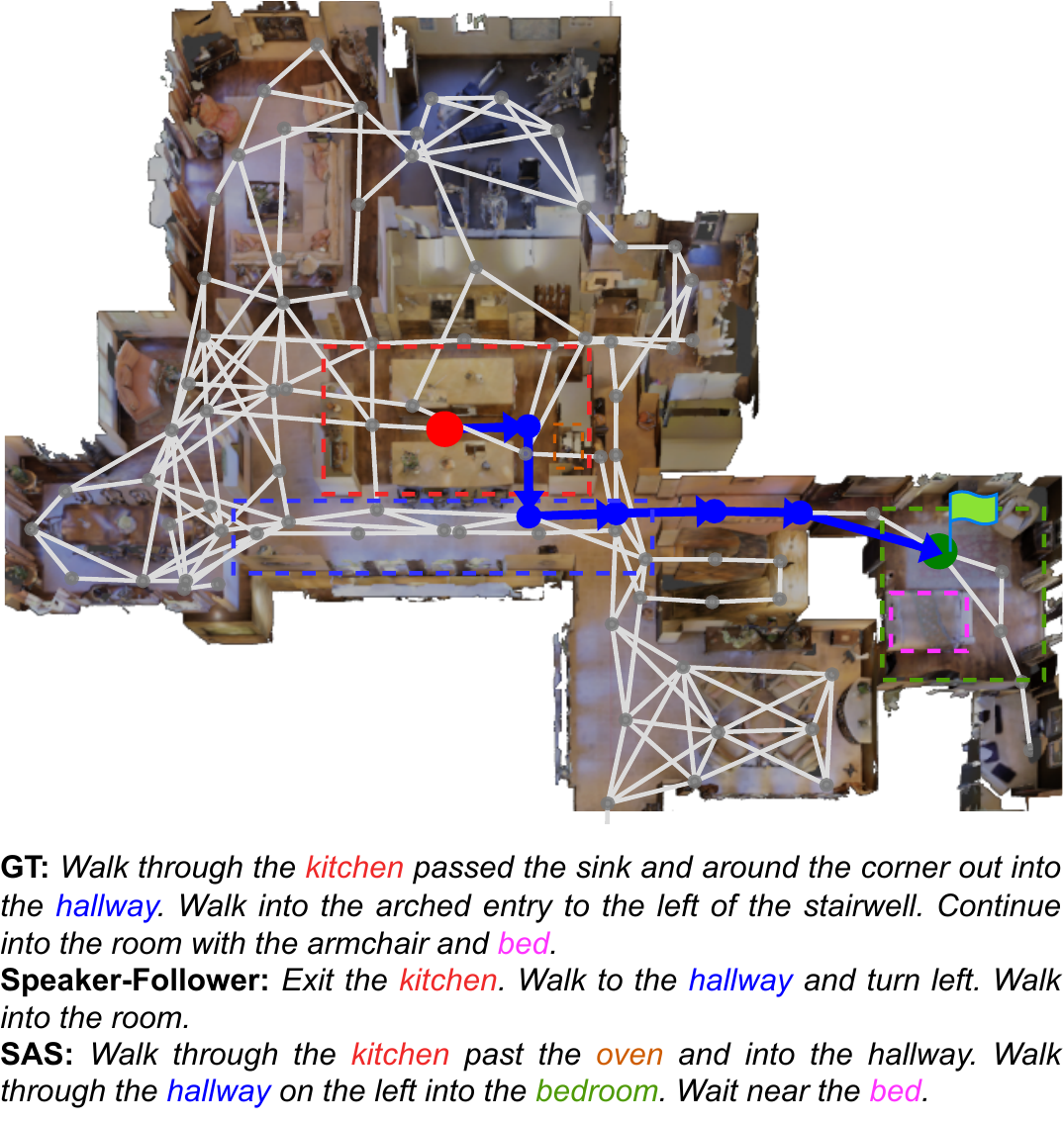}
    \caption{An example of a trajectory and the corresponding generated instruction using SAS$_{ARL+TF}$ model.}
\label{fig:success_r2r_QUCTc6BB5sX173}
\end{figure}
Our SAS model is able to generate meaningful instructions by including object and scene relevant tokens (such as "oven") as shown in Fig. \ref{fig:success_r2r_QUCTc6BB5sX173} that are not referenced in the ground-truth instruction, while the Speaker-Follower baseline model produces shorter and action-focused sentences.

\subsection{Ablation Studies}
\label{sec:ablation_study}
\begin{table}[ht]
    \centering
    \setlength{\tabcolsep}{1.2pt}
\fontsize{7.5}{7}\selectfont
        \caption{Ablation Study (\S\ref{sec:ablation_study}) on R2R dataset \texttt{ValUnseen} split}
    \begin{tabular}{cccccccccccc}
    \toprule
    Met. & PM & TAL & CNN & GRU & SPICE & CIDEr & METEOR & ROUGE & BLEU-4 \\
    \midrule
    \#1 & &&&& 22.5 & 38.0 & 23.9 & 48.3 & 26.2\\
    \#2 & &&&& 22.8 & 39.1 & 23.9 & 49.5 & 27.3\\
    \hline
    \#3 & \checkmark &&&& 21.2 & 39.9 & 23.6 & 50.1 & 27.8\\
    \#4&&\checkmark&&&  22.1 & 40.4 & 24.1 & 53.2 & 28.6\\
    \#5&\checkmark&\checkmark&&& 22.2 & 44.9 & 26.3 & 55.4 & 30.2\\
    \hline
    \#6&\checkmark&\checkmark&\checkmark&& 24.3 & 42.9 & 26.3 & 54.4 & 30.3\\
    \#7&\checkmark&\checkmark&&\checkmark& 24.8 & 43.5 & 25.7 & 56.5 & 33.8\\
    \bottomrule
    \end{tabular}
    \label{tab:abalation-r2r}
\end{table}
Here we ablate on different augmentation and training methods (Table \ref{tab:abalation-r2r}) evaluated on R2R \texttt{ValUnseen} split. Method $\#1$ uses the aforementioned Student forcing (SF) and $\#2$ represents Teacher forcing (TF) for training. When Path Mixing is applied to the SAS model ($\#3$), the Speaker learns the frequent object tokens in the instruction and how to correlate them with the visual features. Models $\#4$ and $\#5$ ($\textsc{SAS}_{TF}$), trained using TAL, learn to associate the object tokens with the actions from the trajectory and inversely co-relate navigational actions with action phrases in the ground truth instructions (\textit{Walk through the double door...}). The models $\#6$ and $\#7$ ($\textsc{SAS}_{ARL+TF}$) are trained to improve the SPICE score using ARL. Using ARL and a GRU-based reward model ($\#7$) has an advantage over using the CNN-based reward model ($\#6$), which produces the highest scores. 

\subsection{Spatial and Semantic effectiveness}
\label{sec:spa_sem_eff}
To study the effectiveness of the proposed spatial and semantic encoding, we measure the amount of object and spatial phrases mentioned in the generated instructions for the speakers evaluated in the R2R \texttt{ValUnseen} environment.

\begin{table}[ht]
    \centering
    \fontsize{9}{10}\selectfont
        \caption{Spatial and Semantic referrals  (\S\ref{sec:spa_sem_eff}) on R2R \texttt{ValUnseen} environment}
    \begin{tabular}{lccc}
    \toprule
    Method & Obj. &  Act. & NonStop  \\
    \midrule
    Human & 14123 (6.01) & 8933 (3.80) & 36689 (15.61)\\ %\hline
    LANA & 2861 (3.62) & 2842 (3.62) & 9685 (12.36) \\
    SAS & 3379 (4.31) & 3184 (4.06) & 10790 (13.78) \\
    \bottomrule
    \end{tabular}
    \label{tab:wordcount-r2r}
\end{table}

In Table \ref{tab:wordcount-r2r}, the values outside the parentheses represent the total counts of objects/landmarks (Obj.) i.e. \textit{chair, bathroom, etc.}, actions/direction phrases (Act.) i.e. \textit{turn left, top of, etc.}, and non-stopwords (NonStop). Meanwhile, the values in parentheses denote the average number of entities per instruction for each respective category. We observed that the SAS speaker includes 18.11\% more objects and landmark entities and 12.03\% more action/direction phrases compared to the LANA speaker. Furthermore, the average length of the instructions is also higher for SAS, indicating richer or more detailed instructions. Although the SAS model did not refer to all the objects or landmarks in the ground truth instructions (SAS: 4.31, Human: 6.01), it includes more action/direction phrases (SAS: 4.06, Human: 3.80). This suggests a better specificity for actions and spatial awareness.

\subsection{Limitations}
\label{sec:limitations}
\begin{figure}[htp]
    \centering
    \includegraphics[width=\columnwidth]{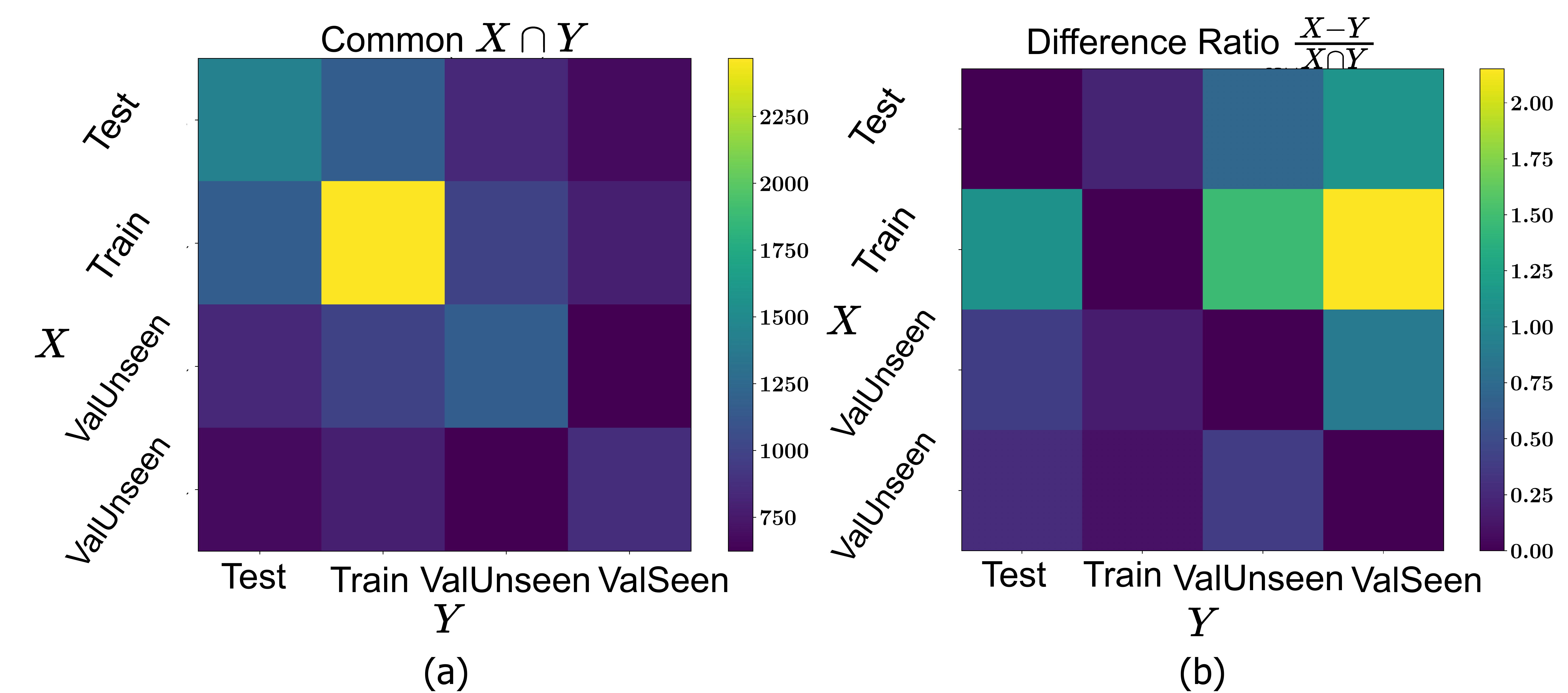}
    \caption{Unique instruction words present in R2R dataset splits. (a) common words between splits, (b) shows the ratio of number of different words to number of common words in between the splits.}
    \label{fig:wordstats}
\end{figure}
Large-scale datasets featuring a broad variety of human-annotated navigation instructions are rare, presenting a significant challenge in the field. Our approach seeks to navigate this obstacle by leveraging a small-scale dataset originally compiled with a different objective: to facilitate the learning of navigation from instructions. It is important to consider this context when evaluating our method's performance, as it operates under the constraint of limited data diversity and volume. Next, we explain some of the challenges in the dataset.

R2R dataset exhibits a notable variation in unique tokens across its different splits: \texttt{Train}, \texttt{ValSeen}, \texttt{ValUnseen}, and \texttt{TestUnseen}. Figure \ref{fig:wordstats} (a) underscores the differences in token commonality across these splits, the disparities being largely attributed to the frequency of tokens in each split. Figure \ref{fig:wordstats} (b) shows the ratio of different words to that of common words in each split, revealing that the \texttt{Train} split, in particular, contains a significant number of unique word tokens compared to common words with respect to other splits. This variation poses a unique challenge for our SAS model, which aims to mimic the instruction distribution of the training set, but may diverge from the linguistic characteristics of the \texttt{ValSeen} and \texttt{ValUnseen} splits, potentially negatively impacting evaluation scores.

\section{Conclusion}
This work proposes a novel navigation instruction generation model that can produce diverse instructions by attending to several structural and semantic cues from the environment. By providing objects, their locations, and their relationships from the scene, Spatially-Aware Speaker can refer to important aspects of the scene in the instruction. An adversarial reward learning method encourages the model to generate diverse instructions. The results show that our method improves on the standard evaluation metrics and performs better than the baselines. 

\paragraph{Future work} In future work, integrating the SAS model with multimodal transformer architectures will be crucial for enhancing the generation of open-vocabulary embodied instructions. This direction promises to overcome the limitations posed by current dataset constraints and improve performance in instruction generation tasks.
% Our SAS model incorporates landmarks, turn-by-turn actions and important objects from the scene into the generated instruction. This not only helps generate more training examples for Embodied Navigation agents, but also can be used to generate instructions for humans for way-finding and guidance.  We believe SAS model will inspire new multi-modal natural language generation (NLG) in the Embodied Navigation domain.

%% file: appendix.tex
\newpage
% \appendix
% \onecolumn
% \begin{center}
\Large
\textbf{Supplementary material for the manuscript titled \textit{"Spatially-Aware Speaker for Vision Language Navigation Instruction Generation"}}
\begin{appendices}
% \end{center}
% \nopagebreak
% \twocolumn
\normalsize	
\section{Adversarial Reward Learning algorithm}
\label{alg:arl}
We use Reward Learning (ARL) extended from  \cite{Wang2018NoMetrics}. We alternatively train SAS policy and reward models for 100 epochs each. The speaker policy is optimised using the loss functions (\S\ref{sec:training}) explained in the main text.   
\begin{algorithm}
\caption{Adversarial Reward Learning Algorithm}
\begin{algorithmic}[0]
\For{epoch $\leftarrow$ 1 to K}
    \State Obtain instruction $X \leftarrow \Pi_\omega$ \Comment{SAS Speaker Policy}
    \If{Train-Policy}
        \State Obtain instruction $\Tilde{X} \leftarrow D$ 
        \State Update Speaker Policy gradient
    \ElsIf{Train-Reward}
        \State Update Reward gradient \Comment{CNN or GRU}
    \EndIf
\EndFor
\end{algorithmic}
\end{algorithm}

\section{Algorithm to extract action phrases from instruction}
\label{alg:action_parsing}
\begin{algorithm}
\caption{Algorithm to extract action phrases}
\begin{algorithmic}[!bth]
\State \textbf{Input:} Navigation Instruction $X$
\State \textbf{Output:} Action phrases $l_{\boldsymbol{X}}$
\State Initialise the empty action phrase list $l_{\boldsymbol{X}}$.
\State $l_{typ}\gets\texttt{\{TRANVERB,DITRANVERB,INTRANVERB,}$
\texttt{NOUN\}}
\State $l_{pos} \gets \texttt{\{ADV,ADP,INTJ,DET,INTRANVERB\}}$
\For {$w_j$, $w_{j+1}$ in $X$}
    \State $c_{w_j} \gets \textbf{check\_verb($w_j$)}$
    \State $c_{w_{j+1}} \gets \textbf{check\_verb($w_{j+1}$)}$
    \If {$c_{w_j}$ in $l_{typ}$}
        \If {$c_{w_{j+1}}$ in $l_{pos}$}
            \State $l_{\boldsymbol{X}} \gets (w_j;w_{j+1})$ \Comment{Join words as a phrase}
        \ElsIf {$c_{w_j} \neq \texttt{NOUN}$}
            \State $l_{\boldsymbol{X}} \gets w_j$ \Comment{Verb as a phrase}
        \EndIf
    \EndIf
\EndFor
\end{algorithmic}
\end{algorithm}

\section{Implementation Details}

We implement our method with PyTorch library. The instruction speaker is trained with AdamW Optimizer \cite{LoshchilovH19} for 100k iterations. Consistent with previous work, panoramic visual features are extracted using a ResNet-152 model, and the angle feature used is 128 dimensional. We retrieve $K = 6$ objects and relationships from Visual Genome for each view direction. The dimension of the hidden state is set to 512. All experiments were performed on an NVIDIA RTX A6000 GPU. $SAS_{TF}$ model is trained to 100K iterations. The $SAS_{TF+ARL}$ model uses pre-trained  $SAS_{TF}$ weights for initialisation. In ARL strategy, policy and reward models are trained alternatively every 100 iterations. All parameters were chosen based on the performance on the \texttt{ValUnseen} split. SAS model has 14.3M learnable parameters. The losses are mixed using the weights $\lambda_{LM} : \lambda_{ULS} : \lambda_{TAL}$ is $2:1:1$. The implementation of the metrics is obtained from the publicly available COCO Caption Evaluation Toolkit \footnote{\url{https://github.com/tylin/coco-caption}}. spaCy package uses the \verb|en_core_web_lg| model.
\subsection{Hyperparameter Search}
The hyper parameters for the final model are selected from the highest SPICE score on the \texttt{ValUnseen} split. The learning rate is fixed based on observing the reward, IRL scores, and supervision losses to prevent model from over-fitting. 

\subsection{Dataset Statistics}
\paragraph{R2R} It has four splits, i.e., \texttt{Train} (61 scenes, 14k instructions), \texttt{ValSeen} (61 scenes, 1k instructions), \texttt{ValUnseen} (11 scenes, 2k instructions), and \texttt{TestUnseen} (18 scenes, 4k instructions). \texttt{Train} and \texttt{ValSeen} splits share house scenes but not with the unseen splits. 
\paragraph{R4R} It extends R2R and contains three sets, i.e., \texttt{Train} (61 scenes, 233k instructions), \texttt{ValSeen} (61 scenes, 1k instructions), and  \texttt{ValUnseen} (11 scenes, 45k instructions).

\section{Qualitative Examples}
\label{app:qualitative_examples}
Here we list some successful instructions generated by the SAS model against the speaker follower \cite{Fried2018SF} baseline. 
\subsection{Successful Example 1}
\textbf{Baseline}: \textit{Walk into the garage and turn right. Turn right and walk into the room. Walk into the bathroom. Stop.} \\
$\textbf{SAS}_{TF}$: \textit{Walk into the garage and turn right. Walk into the room with the two large chairs. Turn right and walk into the bathroom. Wait near the sink.} \\
$\textbf{SAS}_{ARL+TF}$: \textit{Walk through the garage and turn right at the cupboard. Walk into the living room towards the couches. Walk into the bathroom and wait near the sink.} \\
\textbf{GT}: \textit{Go straight into the garage. Turn right and go into the room. Turn right and go into the bathroom. Wait near the sink. }

\subsection{Successful Example 2}
\textbf{Baseline}: \textit{Walk the stairs.} \\
$\textbf{SAS}_{TF}$: \textit{Walk out of the bedroom. turn left and walk up the stairs . stop on the second step from the bottom .} \\
$\textbf{SAS}_{ARL+TF}$: \textit{Walk past the television and out of the bedroom. Turn left and walk up the stairs. Stop on the third step from the bottom.} \\
\textbf{GT}: \textit{Walk through the bedroom and out into the hall way. Turn left and walk up to the stairs. Walk up to the first step and stop. }

\section{More visualisation of trajectories}
\subsection{Successful Example 1}
\begin{figure}[htp!]
    \centering
    \includegraphics[width=\columnwidth]{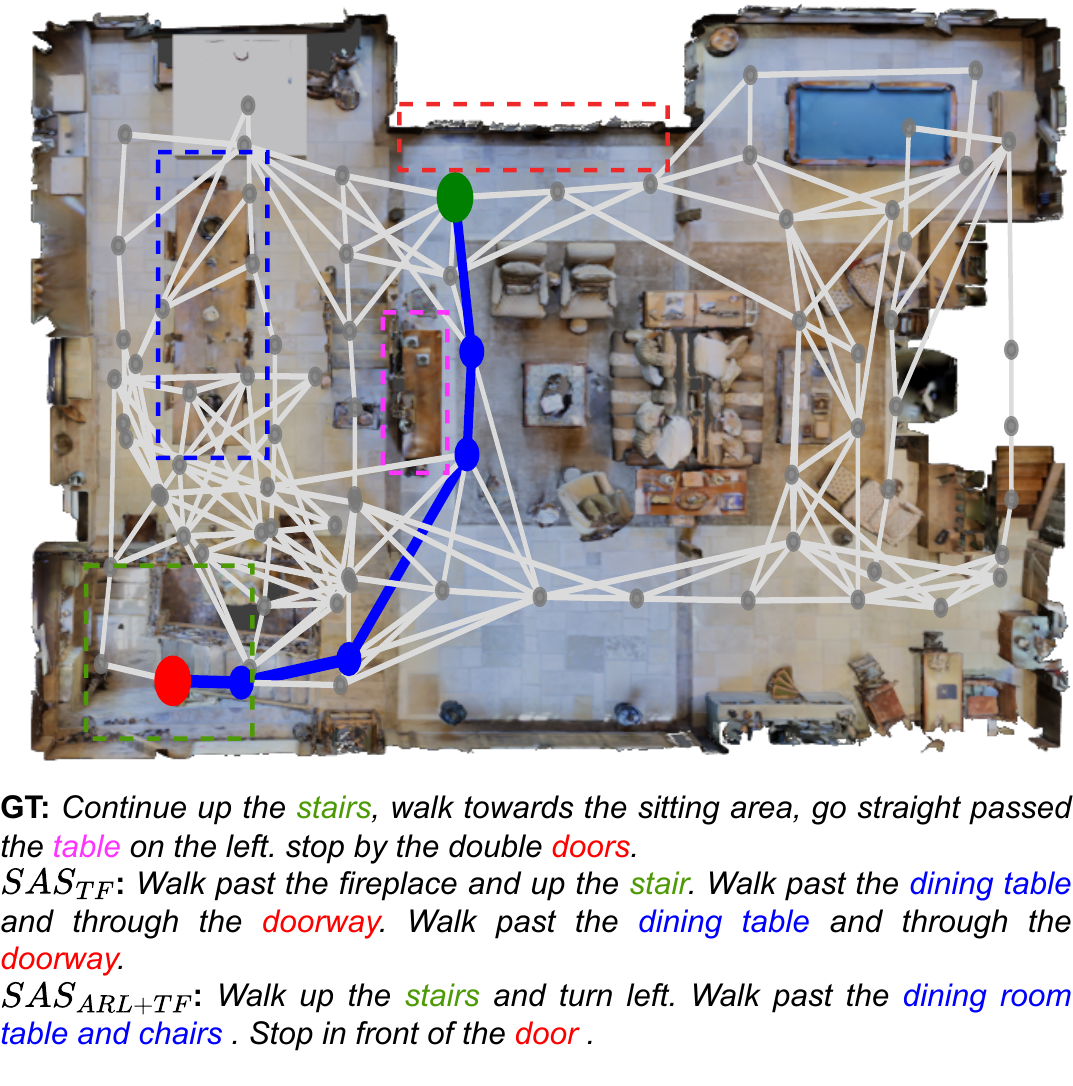}
    \caption{Example of a Successful Instruction Generation.}
    \label{fig:success-instruction}
\end{figure}

Figure \ref{fig:success-instruction} shows one of the success cases. Here, $SAS_{TF}$ generated the correct transition phrases, but failed to mention any stop locations. On the other hand, $SAS_{ARL+TF}$ refers to objects outside of reference sentences and also identifies the correct stop location. This is a good example of Panoramic Room-Object Attention in working.
\subsection{Failure Example 1}
\begin{figure}[htp!]
    \centering
    \includegraphics[width=\columnwidth]{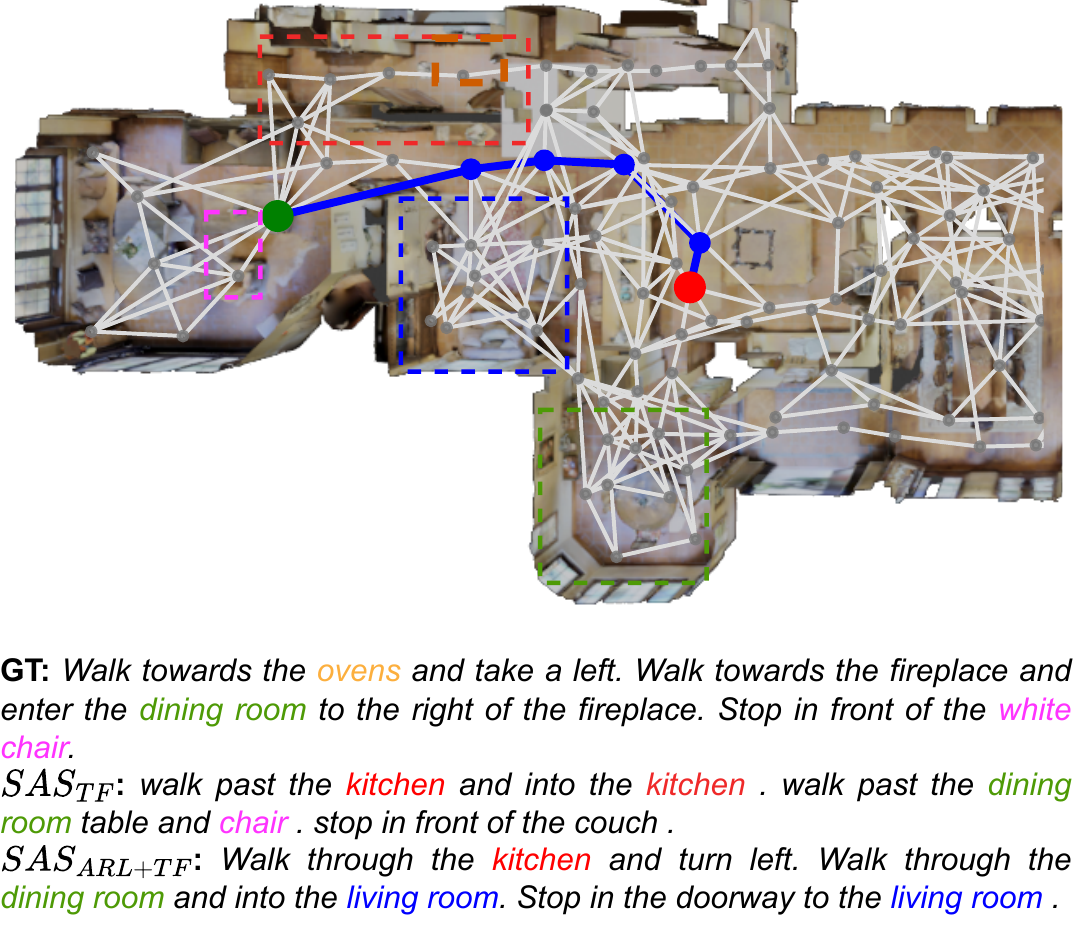}
    \caption{Example of a Failed Instruction Generation.}
    \label{fig:failed-instruction}
\end{figure}

Figure \ref{fig:failed-instruction} shows one of the failure cases. Here both SAS models fail to generate the correct instruction even though the rooms (living room, kitchen, dining room) and objects (table, chair) are identified. The models failed to align the objects, rooms and actions in the correct sequence.

\subsection{Learning Curves}
\begin{figure}[]
    \centering
    \includegraphics[width=\columnwidth]{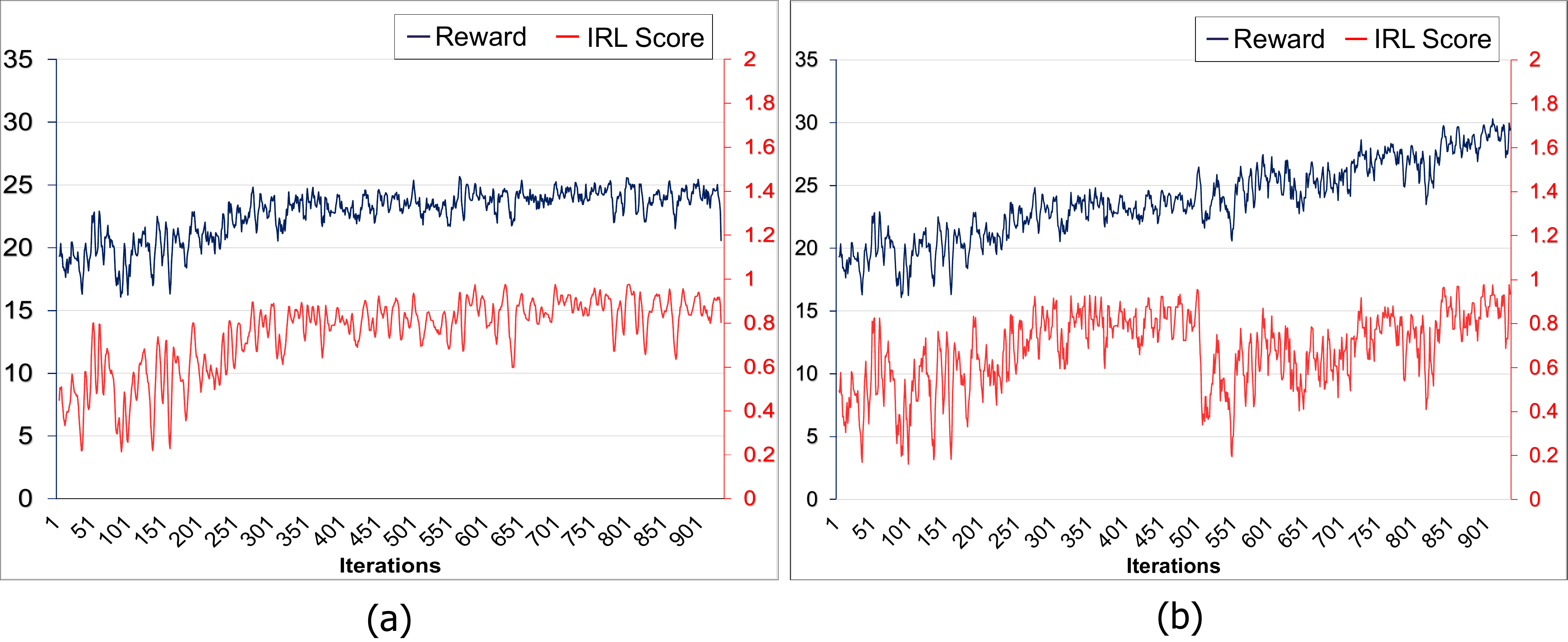}
    \caption{SPICE Scores and Rewards for (a) CNN-based and (b) GRU-based reward models.}
    \label{fig:irlreward}
    % \Description{Two graphs showing SPICE score of the model and predicted reward for two discriminators - CNN and GRU-RNN. The latter learns difference between policy generated and ground-truth instructions to produce better rewards}
\end{figure}

Figure \ref{fig:irlreward} shows the average ARL rewards and SPICE scores from training the model with CNN and GRU reward models. Both models approach a reward close to 1, while the GRU reward model obtains a higher SPICE score. This shows that recurrence can help the discriminator learn the difference between policy-generated and ground-truth instructions. A CNN can extract global information from the instruction but loses temporal information. As temporal information is crucial for navigational instructions and GRU can encode temporal aspects of the instruction as well as visual observations, it produces a higher reward for instructions closer to the ground truth. 

% \end{appendices}

\section{Ethical Considerations}
Embodied navigation stands as a promising frontier with the potential to revolutionise the landscape of language understanding for robots, thus facilitating their seamless integration into everyday human life. However, any effort that involves human-robotic interaction requires a steadfast commitment to upholding ethical, privacy, safety, and legal standards. Although the metrics employed to assess our work align with those commonly used in the machine generation domain, further investigations are imperative to ensure the ethical and safety considerations associated with the instructions generated using automatic methods and their use in the real world.

Our research draws on the R2R \cite{Anderson2018R2R}, FGR2R \cite{Hong2020FGR2R} and R4R \cite{Jain2020R4R} datasets under the MIT licence, which feature an extensive collection of indoor photos captured from American houses, licenced by Matterport3D\footnote{\url{https://kaldir.vc.in.tum.de/matterport/MP_TOS.pdf}}. The Matterport3DSimulator, used in our experiments, is also under MIT license. To protect privacy and confidentiality, the providers of the datasets have anonymised both the houses and the associated photos. In addition, the navigational instructions derived from these datasets are devoid of explicit language. As a result, our work shows minimal ethical, privacy, or safety concerns.

\end{appendices}